%% file: aaai26.tex
\documentclass[letterpaper]{article} 
\usepackage{aaai2026}  
\usepackage{times}  
\usepackage{helvet}  
\usepackage{courier}  
\usepackage[hyphens]{url}  
\usepackage{graphicx} 
\usepackage{natbib}  
\usepackage{caption} 
\usepackage{algorithm}
\usepackage{algorithmic}
\usepackage{booktabs} 
\usepackage{amsmath} 
\usepackage{amsfonts}

\usepackage{newfloat}
\usepackage{listings}
\DeclareCaptionStyle{ruled}{labelfont=normalfont,labelsep=colon,strut=off} 
\floatstyle{ruled}
\newfloat{listing}{tb}{lst}{}
\floatname{listing}{Listing}
\title{Title}

\title{Towards Reinforcement Learning from Neural Feedback: Mapping \\ fNIRS Signals to Agent Performance}

\author{
    Julia Santaniello\textsuperscript{\rm 1},
    Matthew Russell\textsuperscript{\rm 1},
    Benson Jiang\textsuperscript{\rm 1},
    Donatello Sassaroli\textsuperscript{\rm 1},
    Robert Jacob\textsuperscript{\rm 1},
    Jivko Sinapov\textsuperscript{\rm 1}
}

\affiliations{
    \textsuperscript{\rm 1}Tufts University, Department of Computer Science\\
    \{julia.santaniello, benson.jiang, donatello.sassaroli, jivko.sinapov\}@tufts.edu, \{mrussell, jacob\}@cs.tufts.edu
}

\usepackage{bibentry}

\begin{document}

\maketitle

\begin{abstract}
Reinforcement Learning from Human Feedback (RLHF) is a methodology that aligns agent behavior with human preferences by integrating user feedback into the agent's training process. This paper introduces a framework that guides agent training through implicit neural signals, with a focus on the neural classification problem. Our work presents and releases a novel dataset of functional near-infrared spectroscopy (fNIRS) recordings collected from 25 human participants across three domains: Pick-and-Place Robot, Lunar Lander, and Flappy Bird. We train multiple classifiers to predict varying levels of agent performance (optimal, suboptimal, or worst-case) from windows of preprocessed fNIRS features, achieving an average F1 score of 67\% for binary and 46\% for multi-class classification across conditions and domains. We also train multiple regressors to predict the degree of deviation between an agent's chosen action and a set of near-optimal policy actions, providing a continuous measure of performance. Finally, we evaluate cross-subject generalization and show that fine-tuning pre-trained models with a small sample of subject-specific data increases average F1 scores by 17\% and 41\% for binary and multi-class models, respectively. Our results demonstrate that mapping implicit fNIRS signals to agent performance is feasible and can be improved, laying the foundation for future Reinforcement Learning from \textit{Neural} Feedback (RLNF) systems.
\end{abstract}

\begin{links}
    \link{Dataset}{https://github.com/your-profile/fNIRS2RL}
    \link{Code}{https://github.com/your-profile/NeuroLoop-Classification/tree/aaai26}
\end{links}
\input{Sections/introduction}
\input{Sections/background}
\input{Sections/dataset}
\input{Sections/methods}
\input{Sections/machinelearning}

\input{Sections/results}
\input{Sections/conclusion}
\section{Acknowledgments}
We are grateful for the support and feedback of the MuLIP and HCI labs at Tufts University, especially Kenny Zheng, Anes Kim, Anna Sheaffer, Brennan Miller-Klugman, and Iris Yang.
\bibliography{references}
\end{document}

%% file: Sections/introduction.tex
\section{Introduction}
Recent advances in Reinforcement Learning from Human Feedback (RLHF) have become crucial for training and fine-tuning state-of-the-art systems~\cite{wu2023finegrained,mosqueiraSurvey}. Specifically, RLHF has seen significant success by addressing limitations that plague common Reinforcement Learning (RL) techniques. Integrating human feedback has been incredibly beneficial for aligning agent behavior with human preferences. However, many feedback methods require active participation and/or expert demonstration for evaluative feedback to be most effective \cite{christiano, liRLsurveyRLHF, retzlaff2024human}. In addition, most approaches are limited to explicit feedback, which may lack insight into the nuances of human decision-making and internal assessments. Techniques that retrieve richer feedback may require tasks that impose greater cognitive effort on users \cite{lindner2022humans}. These drawbacks can be inconvenient for the evaluator or limit accessibility to certain user groups. They may also contribute to higher mental workload and unnatural, shallow feedback \cite{casper2023open}. These limitations may make it difficult to develop adaptive AI technologies that fully align with human preferences while being available to all user groups. 
%
\begin{figure}[t!]
\centering
\includegraphics[width=8.5cm]{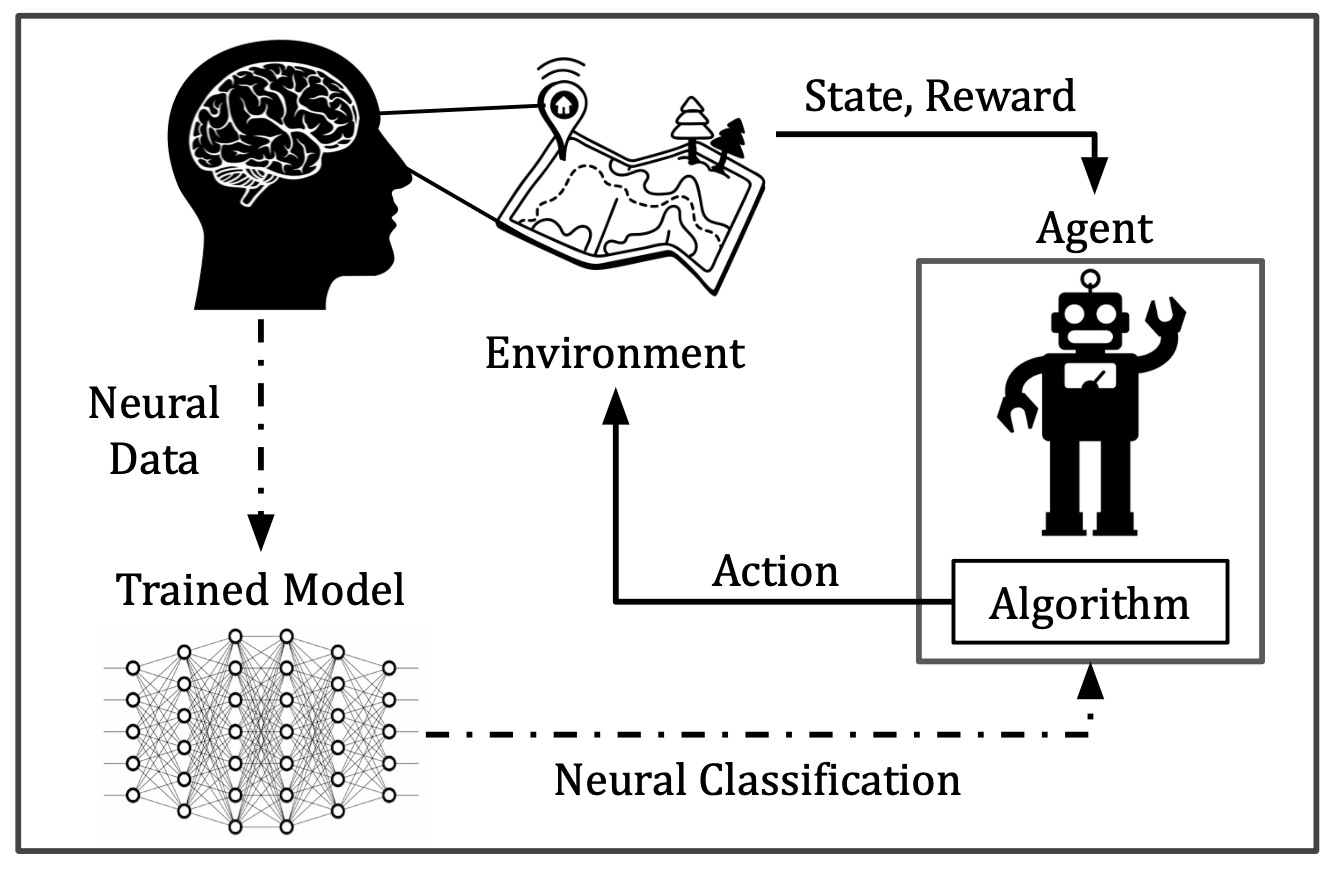}
\caption{NEURO-LOOP: A high-level diagram of the proposed framework. This paper addresses the neural classification problem with the intent to apply trained models to the proposed pipeline.}
\label{fig:neurloop-diagram}
\end{figure}

\textit{Implicit} human feedback systems present a valuable alternative by allowing agents to learn from natural human responses. Modern methods focus on facial expression or gesture classification to adapt agent behavior at little to no additional cognitive cost to the teacher \cite{cui2021empathic}. However, some of these procedures still require explicit instruction and conscious physical effort to adjust gestures and expressions. Further, reliance on private training data poses challenges for data collection and distribution.

To address these limitations, we propose \emph{NEURO-LOOP}, a fully implicit neural feedback framework that leverages fNIRS signals to directly guide agents' training using RLHF techniques. This framework employs passive Brain-Computer Interfaces (BCI) to align agent behavior with user intent through passive observation. We employ a functional near-infrared spectroscopy (fNIRS) device to record brain activity. fNIRS is a non-invasive neuro-imaging tool that can track sustained cognitive states such as mental workload, attention, and decision-making. This technology measures the hemodynamic response, or change in blood flow, in the prefrontal cortex (PFC) over time.

This paper addresses the neural classification problem as the first step toward integrating implicit fNIRS signals into RLHF frameworks. Our contributions are outlined below:

\begin{itemize}
    \item We design and implement a controlled experimental protocol to investigate whether functional near-infrared spectroscopy (fNIRS) signals reflect varying levels of agent performance during human-agent interactions. We publicly release a novel dataset comprised of synchronized fNIRS recordings and agent transition variables.
    
    \item We demonstrate that machine learning can distinguish multiple levels of agent performance, extending past binary classification and enabling finer-grained feedback while maintaining low cognitive workload for users.
    
    \item We evaluate model performance and generalization between participants, suggesting that cross-subject transfer remains a challenge, but subject-specific calibration using limited data is feasible and improves performance.
\end{itemize}
We further compare model performance and user workload across both \textit{passive} (observation) and \textit{active} (physical demonstration) tasks, establishing a benchmark for future research in Reinforcement Learning from \textit{Neural} Feedback (RLNF).

%% file: Sections/background.tex
\section{Related Work}
Reinforcement Learning from Human Feedback (RLHF) allows agents to learn from human preferences through a range of approaches. These methods can be described as having either implicit or explicit feedback mechanisms.
\\\\
\textbf{Explicit Feedback:} Explicit feedback is the most common modality used to shape an agent's policy to align with human expectations. Methods in this category often provide binary or scalar evaluative signals, such as preference labels or human-provided rewards, to shape the agent's policy or learning process \cite{griffith,KnoxStone2011AugmentingRL}. Past research indicates that evaluative metrics with wider ranges have greater training success \cite{aabl}. These interactions are often shorter-term due to increased cognitive workload and sustained physical demand. Imitation learning is a related approach that often requires expert demonstration or active participation for the agent to learn from user examples effectively \cite{imitation}.
\\
\textbf{Implicit Feedback:} Implicit feedback leverages unobtrusive human responses to enhance reinforcement learning agent behavior. These feedback techniques often include gaze tracking \cite{Veeriah2016FaceValuing}, expression recognition \cite{Arakawa2018DQNTAMERHR}, or gesture classification \cite{cui2021empathic} to shape agent policies. One drawback is that some of these procedures require explicit instruction, as participants must consciously and physically adjust their gestures or expressions to provide meaningful feedback. We propose using neural signals to communicate evaluative feedback directly with minimal instruction.
\begin{figure}[t] 
    \centering
    \begin{minipage}{0.48\linewidth} 
        \centering
        \includegraphics[width=\linewidth]{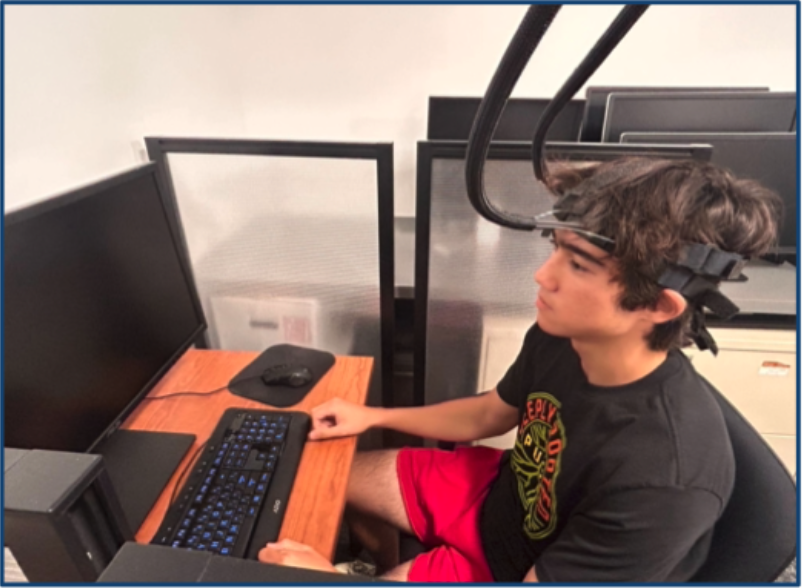}
    \end{minipage}
    \hspace{5pt} 
    \begin{minipage}{0.48\linewidth}
        \centering
        \includegraphics[width=\linewidth]{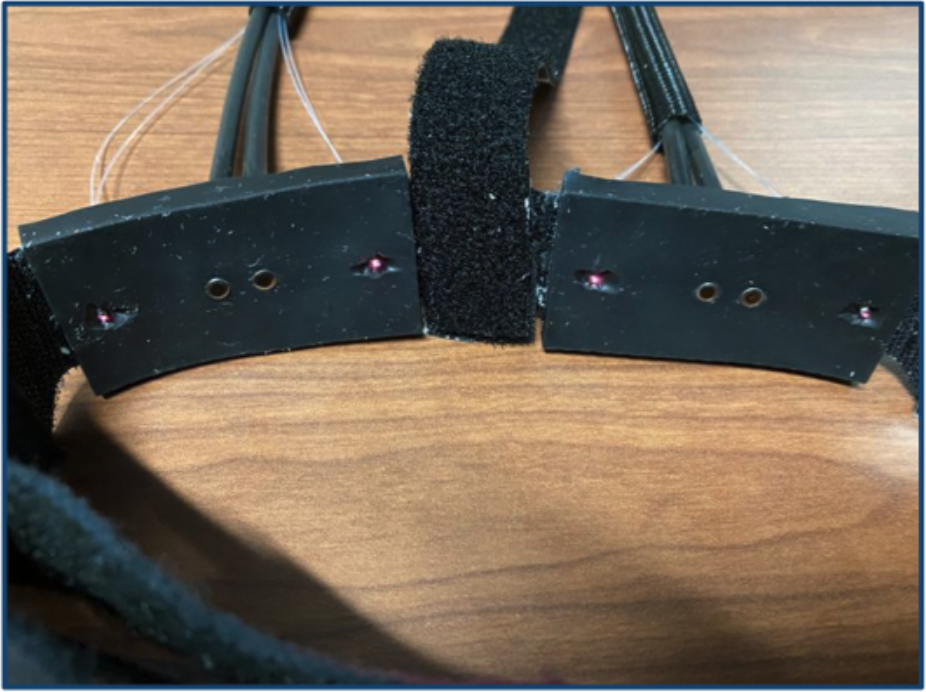}
    \end{minipage}
    \caption{Setup: Participants sat 24 inches in front of a computer screen. The fNIRS device is a headband that shines pulsating infrared light into the PFC to detect changes in blood flow.}
    \label{fig:setup_device}
\end{figure}
\begin{figure*}[ht]
    \centering
    \includegraphics[width=13cm]{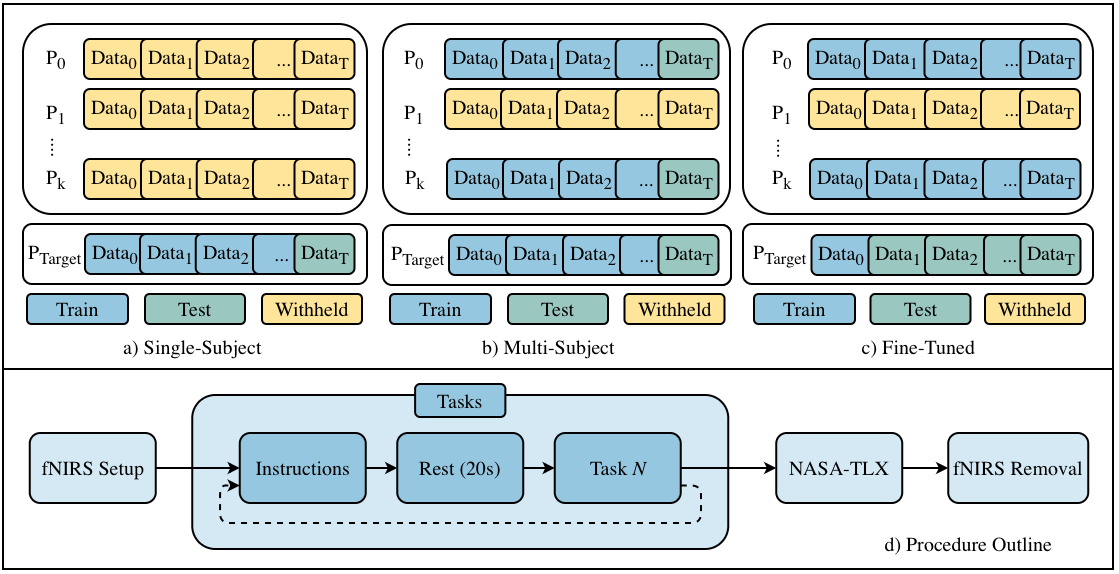}
    \caption{Training Paradigms: Diagram of the three training paradigms and Experimental Protocol. (a) Single-subject models are trained on a set of data from one participant and are evaluated using withheld data from the same participant. (b) Multi-subject models are trained on a set of participants and are evaluated using withheld data from the same set. (c) Fine-tuned models are multi-subject models calibrated with a fraction of a target participant's data.
    }
    \label{fig:paradigms}
\end{figure*}
\\
\textbf{ErrPs for Human-Robot Interaction:}  
Error-related potentials (ErrPs) are neural signals that occur when the brain perceives an error. These signals, recorded via EEG, have been explored as a method for improving human-robot interaction. Previous research has applied ErrP signals to intervene in robotic decision-making by providing a binary indicator of error or dissatisfaction. Some work has even integrated ErrPs into RLHF algorithms, demonstrating improved training efficiency~\cite{vukelic_combining_2023,agarwal_humanloop_2020,xu_accelerating_2021}. EEG signals are often transient and susceptible to artifacts \cite{rihet_robot_2024}. We explore the use of fNIRS as a complementary signal for brain-driven RLHF systems. Compared to other neuro-imaging devices, fNIRS produces signals that offer more tolerance to physical movement, greater portability, and higher spatial resolution \cite{Pinti2018Noise}. Using the fNIRS system described by Giles \cite{giles}, we collected phase and intensity signals, making fNIRS suitable for long-horizon naturalistic settings outside the laboratory.
\\
\textbf{fNIRS for Human-Centered Technologies:}  
Recent advancements have made it possible for brain signals to be used in human-centered technologies \cite{humancenteredfnirs}. When applied to adaptive frameworks, fNIRS data often monitors cognitive workload, allowing systems to adjust to user strain. Past work has explored applications like unmanned aerial vehicles (UAV) \cite{afergan_dynamic_2014}, and human-robot systems \cite{solovey_brainput_2012, roy_how_2020}.
\\
\textbf{Motivation:} fNIRS is quite under-explored in reinforcement learning and human-in-the-loop machine learning. fNIRS also offers beneficial characteristics and features that EEG generally lacks. Prior research has correlated fNIRS signals to reward-based gameplay events like video game performance \cite{VideoGamefNIRS2018} and physical game performance \cite{fNIRSRewardDanceSimulation2012}. Other work links the PFC to reward-based decision-making during vicarious and personal video game play \cite{MorelliVicariousPersonal2015}, suggesting hidden potential for alignment with agent evaluation metrics. Therefore, we propose mapping fNIRS signals to various degrees of agent performance, a first step toward creating an fNIRS-driven RLHF system.

%% file: Sections/dataset.tex
\section{Dataset Notation} 
We formalize the structure of the complete dataset as $\mathcal{D}$, and the subset of data from each participant $k$, denoted as $\mathcal{D}^k$. Dataset $\mathcal{D}^k$ is further divided into two subsets, the neural dataset and the task dataset.\\
\textbf{Neural Dataset:}  We define the neural dataset $\mathcal{N}^k$ as the set of neural channel recordings over time for some participant $k$. Let $\mathcal{N}_i \in \mathbb{R}^{M \times T}$ denote the neural signal matrix at instance $i$, where $M$ is the number of neural channels and $T$ is the number of timestamps for some participant. A single neural data vector at timestamp $t$ across all channels can be denoted as $\mathbf{n}_t \in \mathbb{R}^M$. The signal at a specific timestamp $t$ and channel $m$ can be expressed as $n_{t,m}$, where $t \in \{0, \dots, T\}$ and $m \in \{0, \dots, M\}$.
\\
\textbf{Task Dataset:}  We define the Task dataset $\mathcal{H}^k$ as the set of agent transition variables, or learning task statistics, over time for some participant $k$. Let $\mathcal{H}_i \in \mathbb{R}^{P \times T}$ represent the task data matrix at instance $i$, where $P$ is the number of task variables and $T$ is the number of timestamps.
We may describe a vector of learning task statistics as 
$
\mathbf{h}_k = \{S_t, A_t, R_t, S_{t+1}, V_{t}, B_{t}, E_{t}\}_{t=0}^{t=T},
$
where $S_t$ is the agent's state at time $t$, $A_t$ is the action chosen by the agent or human, $R_t$ is the reward, $S_{t+1}$ is the next state, and $V_{t}, B_{t}, E_{t}$ represent various agent performance values. An instance of a task statistic data point at some timestamp for a specific task variable can be denoted as $h_{t,p} = f(H_p, T_t)$, where $p \in \{0, \dots, P\}$.
After the raw neural data has been pre-processed, we hypothesize that a relationship exists between features of the neural data and the learning task statistics outlined above, where $x$ is a neural feature vector. We let $
\hat{y} = \phi(x_i, h_i)$ denote the relationship between a vector of human neural data features at some instance $i$ and its agent performance label.

%% file: Sections/methods.tex
\section{Methodology}
\textbf{Participants:} 
We recruited 25 participants for a mixed within and between-participants study. Participants were between 19 to 27 years old and were recruited through physical and virtual flyers. Fourteen participants identified as female and eleven as male. Each participant completed 3-4 conditions out of a possible six conditions, resulting in at least 10 participants per condition.
\\
\textbf{Equipment:}  
Neural data was collected using an ISS OxiplexTS fNIRS device. This device uses pulsating infrared lasers to calculate the change in hemodynamic response under the human skull. Three OpenAI Gymnasium reinforcement learning domains were adjusted for seamless interfacing with participants: Robot Fetch and Place, Lunar Lander, and Flappy Bird.
%
\\
\textbf{Domains:}  
An agent $\mathcal{A}$ operates in one of three OpenAI Gym domains. In Lunar Lander, it uses \texttt{up}, \texttt{left}, \texttt{right}, and \texttt{down} to land between a set of flags without crashing. Flappy Bird uses actions \texttt{up} and \texttt{down} to clear a sequence of pipes/obstacles. In Robot Fetch and Place, the agent uses \texttt{x-left}, \texttt{x-right}, \texttt{y-up}, \texttt{y-down}, \texttt{z-up}, \texttt{z-down}, and \texttt{gripper-close} to place a cube at a marked goal.
\\
\textbf{High-Level Experimental Procedure:}  
A complete experimental procedure included the administration of an informed consent form, configuration of the fNIRS device, task instruction, and two post-experiment questionnaires (Figure \ref{fig:paradigms}). 
\begin{figure*}[ht]
\centering
\includegraphics[width=14cm]{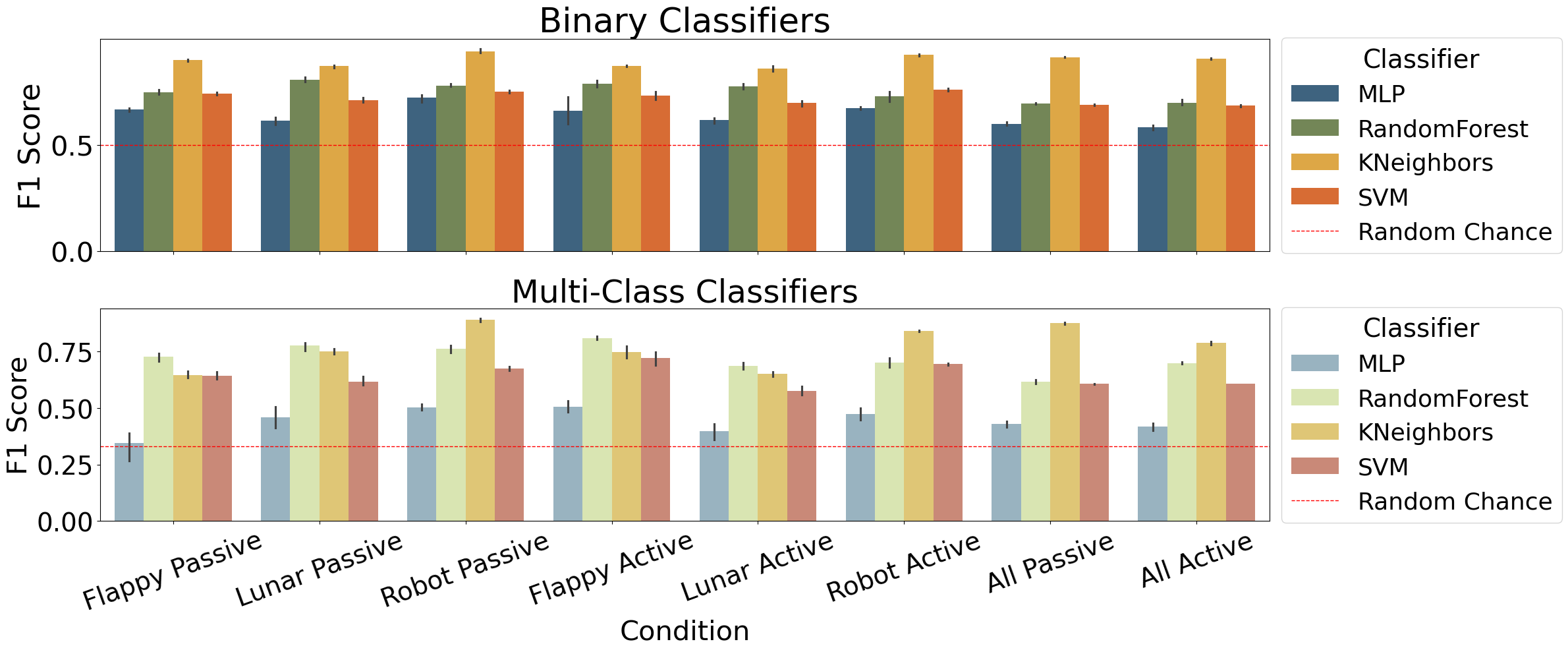}
\caption{Multi-Subject Classification Performance: This figure illustrates classification performance (F1) for binary and multi-class models. Binary models attempted to classify optimal vs. suboptimal behavior, while multi-class models attempted to differentiate between optimal, suboptimal or worst-case agent behavior.}
\label{fig:multi-subject-classes}
\end{figure*}
Each task consisted of a set of episodes in which an agent attempts to complete a domain-specific goal. The participant’s interaction with the agent was either \textit{passive} or \textit{active}. A condition is any combination of domain and interaction. Participants completed 3-4 conditions that lasted between 2-5 minutes each. Conditions were randomly selected, but passive tasks were completed first. Participants were seated 24 inches from the computer screen that displayed the agent and domain. A 20-second rest/pause was taken before each task to calibrate the fNIRS device. The participant completed a NASA-TLX questionnaire for each task, and a post-task questionnaire at the end of the study.
\\
\textbf{Passive Task:}  
The \textit{passive} condition instructed participants to observe and reflect on the performance of an autonomous agent completing a goal within its environment. The autonomous agent initially selects actions from a near-optimal policy and can be described as successful. With some probability $p$, the agent transitions to a non-optimal action selection state and becomes unsuccessful for the remainder of an episode. We refer to this timestep as the \textit{point of failure}. The \textit{degree of failure} labels all actions after the point of failure accordingly. Due to the variation in each domain, the \textit{point of failure} and \textit{degree of failure} are determined differently for each.
\\
In an environment with a \textit{discrete} action space (Lunar Lander, Flappy Bird), non-optimal actions were selected by altering the agent's chosen action distribution. We let \( \mathbf{p} = [p_1, p_2, \dots, p_n] \) be the probability distribution over \( n \) actions. Let \( b \) be the index of the best action (i.e., the action with the highest probability): $b = \arg\max \left( p_i\right)$. Let \( w \) be the index of some non-optimal action (i.e., an action without the highest probability): $w = \arg\min \left( p_i \right)$. The new probability distribution before normalizing \( \mathbf{p'} \) is defined as:
\[
p'_i =
\begin{cases}
0 & \text{if } i = b \\
p_w + p_b & \text{if } i = w \\
p_i & \text{otherwise}
\end{cases}
\]
Then the array is normalized by dividing by the sum. Increasing the value of the lowest probability increases the likelihood that the worst-case and non-optimal actions are chosen.

Optimal Lunar Lander agents often landed between the flags without crashing. Suboptimal agents often landed outside of the goal location. Worst-case agents crashed. Optimal Flappy Bird agents often completed long episodes (15+ pipes cleared). Suboptimal Flappy agents completed medium-length episodes (5-15 pipes). Worst-case Flappy agents often completed short episodes ($\leq$ 5 pipes).

In an environment with a \textit{continuous} action space (Robot Fetch and Place), suboptimal and worst-case action selection was implemented differently. Suboptimal actions altered the robot's goal, resulting in fluid, swift movements to the wrong goal state. Worst-case actions did not change the robot's goal, but added random variance to joint positions, resulting in unnatural, abrupt movements or dropping/throwing the block.

The agent continued to navigate through various action-selection states until the end of the task. The neural data, learning task statistics, and timestamps were stored in a hash map and saved as a de-identified demonstration.
\begin{figure}[b] 
    \centering
    \begin{minipage}{\linewidth}
        \includegraphics[width=0.3\linewidth]{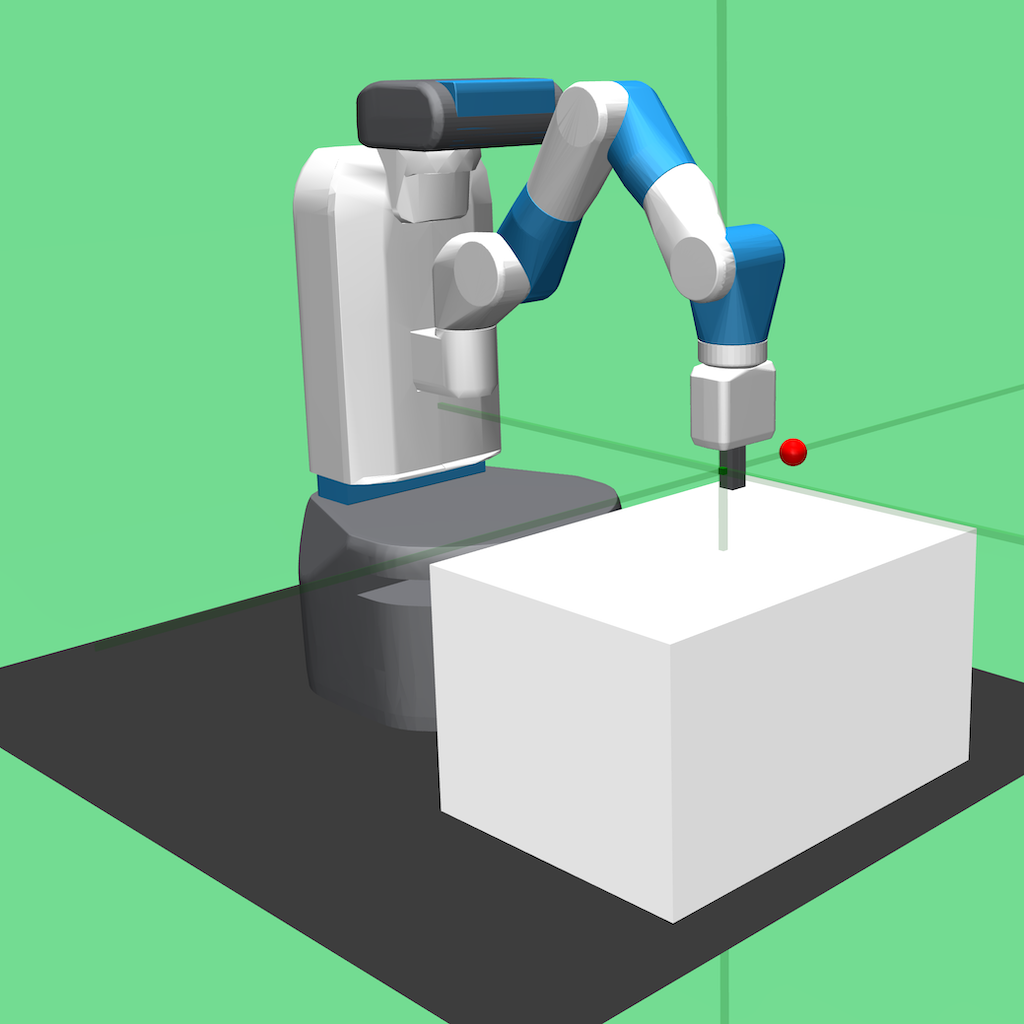} \hspace{2pt}
        \includegraphics[width=0.3\linewidth]{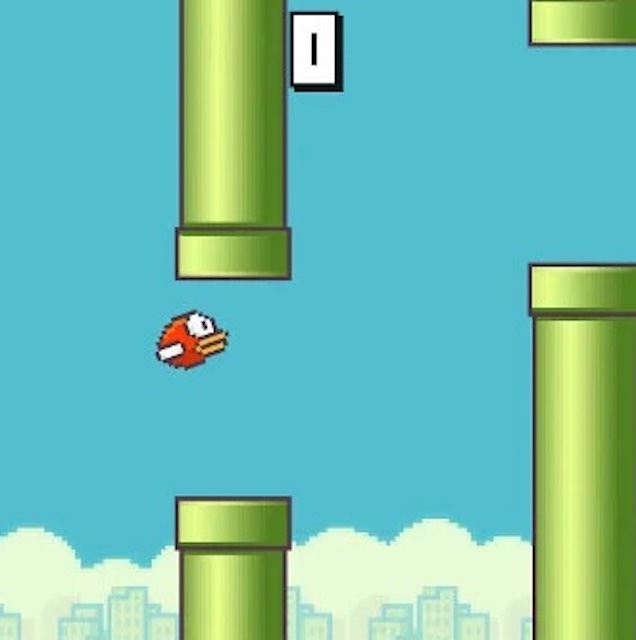} \hspace{2pt}
        \includegraphics[width=0.3\linewidth]{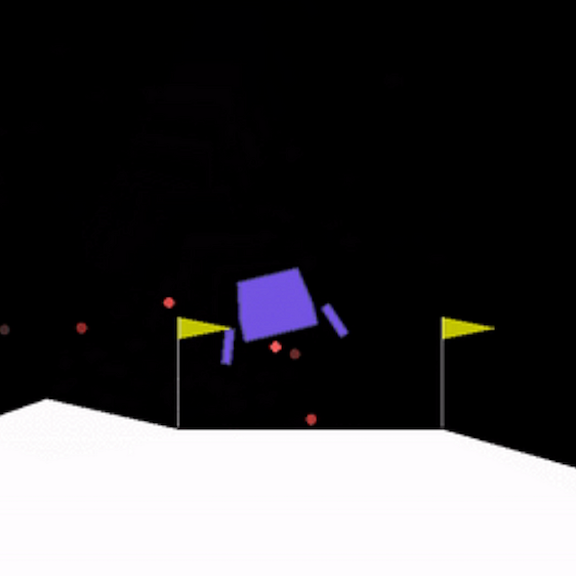}
    \end{minipage}
\caption{Agents and Domains: Robot Fetch and Place, Flappy Bird and Lunar Lander, respectively.}\label{fig:domains}
\end{figure}
\\
\textbf{Active Task:}  
The \textit{active} condition instructed a participant to physically guide an agent towards its goal using a keyboard or joystick. Participants used a basic computer keyboard to guide Flappy Bird and Lunar Lander agents, and an Xbox Controller to guide the Robot arm. As they played, the participant's chosen actions, earned rewards, and seen states were saved. The neural and task data were labeled and saved identically to Passive Tasks.
%
\\
\textbf{Post-Experiment:} 
After the tasks were completed and the fNIRS device was removed from the participant, a NASA-TLX and Post-Task Questionnaire were administered. The NASA-TLX quantifies mental workload through a self-report questionnaire, offering insight into the participant's experience with each condition. The post-task questionnaire offers participants the opportunity to self-report variables that might confound results such as caffeine intake, sleep or technology familiarity.
\begin{figure}[b!]
\includegraphics[width=8cm]{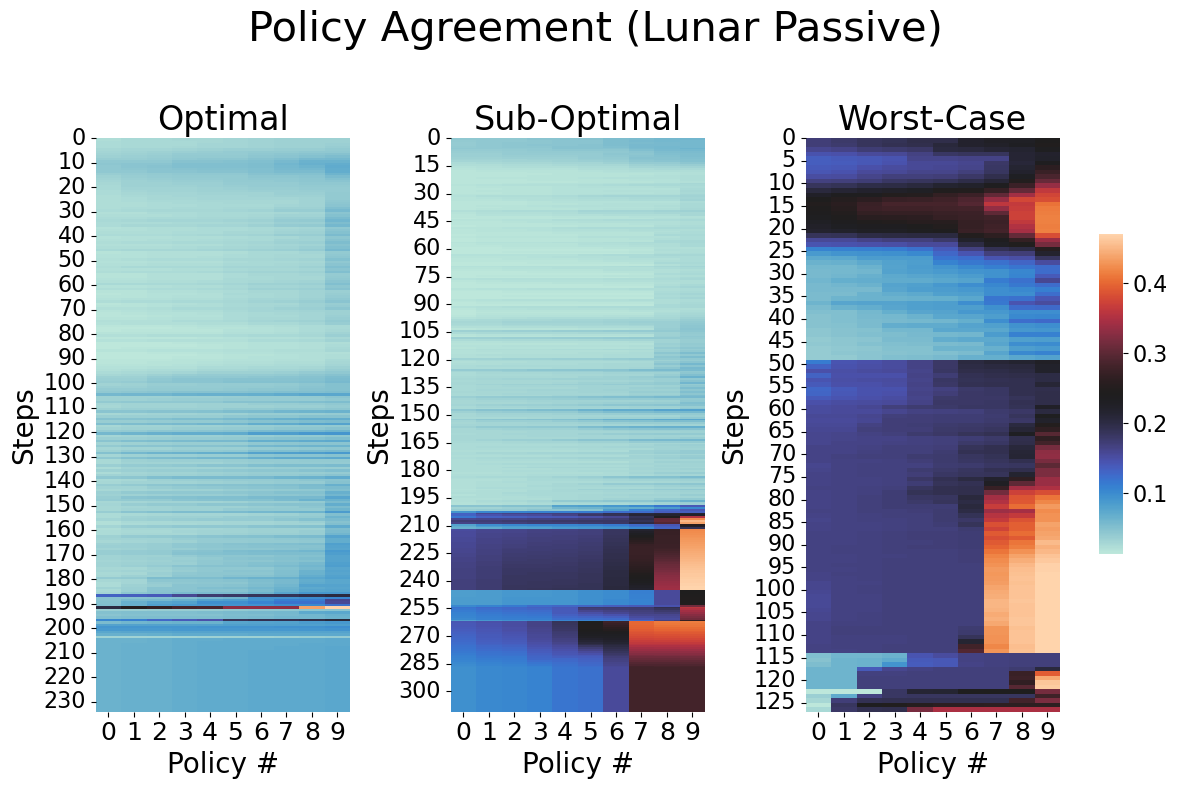}
\centering
\caption{Multi-Policy Agreement Heatmap: These three heatmaps visually illustrate the \textit{point of failure} and the \textit{degree of failure} within each optimality class. Cool colors indicate low error, while warm colors indicate high error.}
\label{fig:heatmap}
\end{figure}
\\
\textbf{Participant Data Summary:} 
Two participants were excluded from data analysis due to technical difficulties or non-consent to the release of their neural data. Participants with inefficient data are also included in the dataset so researchers can explore the effect of noisy fNIRS data for future analyses \cite{suboptData2024}. A Participant Data Summary is included with the dataset outlining eligibility and noteworthy disturbances during data collection.

%% file: Sections/machinelearning.tex
\section{Machine Learning Approaches}
\textbf{Classification Labels:} To train classifiers and regressors, both discrete and continuous labels were recorded. These labels are equivalent to the calculated \emph{degree of failure} and are assigned to all actions after the \emph{point of failure} in an episode. Binary performance labels are denoted as $B_t \in \{0,1\}$, where $0$ represents optimal behavior and $1$ represents suboptimal behavior. Multi-class performance labels are denoted as $\mathbf{V}$. At some time step $t$, $\mathbf{V}_t$ may be some discrete number from the set $ V_t \in \{0,1,2\} $ where $0$ represents optimal, $1$ represents suboptimal, and $2$ represents worst-case behavior. Continuous performance errors are denoted as $E(t) \in \mathbb{R}$. At some time step $t$, $\mathbf{E}_t$ may be some continuous value where lower values indicate near-optimal actions and higher values indicate suboptimal to worst-case actions. This error value was designated by the multi-policy agreement system outlined below.
\\
\textbf{Multi-Policy Action Agreement:}
To calculate \textit{continuous} performance labels, we designed a multi-policy action classification system. In any reinforcement learning problem, there may be more than one optimal path to the same goal. To avoid mislabeling agent behavior, we compare the agent's chosen action with $K$ near-optimal policies, where $K=10$. An error value is calculated between the agent's chosen action and each near-optimal policy's action. Then the average is calculated across the $K$ policies. The error between the agent's discrete action and the \( k \)-th near-optimal policy $\pi_k \left( s_t \right)$ is calculated using Kullback-Leibler (KL) Divergence:
\[
E_k(t) = D_{\text{KL}} \left( a_t \; \| \; \pi_k(s_t) \right) = \sum_{i=1}^n a_{t,i} \log \left( \frac{a_{t,i}}{\pi_k(s_{t})_i} \right)
\]
The error between the agent's continuous action and the \( k \)-th near-optimal policy $\pi_k \left( s_t \right)$ is calculated using Euclidean distance, where $n$ is the size of the action space:
\[
E_k \left( \pi_k \left( s_t \right), a_t \right) = \sqrt{\sum_{i=1}^{n} (\pi_k\left( s_t \right)_i - a_{t,i})^2}
\]
We let \(a_t\) denote the agent's chosen action at time \(t\), and let \( \pi_k(s_t) \) represent the action taken by the \( k \)-th near-optimal policy at time \( t \). The error between the agent's continuous action and the \( k \)-th near-optimal policy $\pi_k \left( s_t \right)$ is defined as \( E_k(t) \). Then, the average error \( \bar{E}(t) \) over \( K \) near-optimal policies is calculated and used as a continuous label.
\begin{table*}[ht]
\centering
\small
\setlength{\tabcolsep}{5pt}
\begin{tabular}{l@{\hskip 10pt}ccc@{\hskip 10pt}ccc}
    \toprule
    \textbf{Condition} & \multicolumn{3}{c}{\textbf{Binary Classification}} & \multicolumn{3}{c}{\textbf{Multi-Class Classification}} \\
    & \textbf{Multi-Subject} & \textbf{Cross-Subject} & \textbf{Fine-Tuned} & \textbf{Multi-Subject} & \textbf{Cross-Subject} & \textbf{Fine-Tuned} \\
    \midrule
    Robot Passive        & 0.72 ± 0.005 & 0.54 ± 0.01 & 0.57 ± 0.03 & 0.50 ± 0.005 & 0.33 ± 0.01 & 0.41 ± 0.02 \\
    Robot Active         & 0.67 ± 0.005 & 0.53 ± 0.04 & 0.56 ± 0.02 & 0.47 ± 0.01 & 0.29 ± 0.02 &  0.35 ± 0.03 \\
    Lunar Passive        & 0.61 ± 0.01 & 0.45 ± 0.03 & 0.56 ± 0.02 & 0.46 ± 0.03 & 0.26 ± 0.03 & 0.36 ± 0.02 \\
    Lunar Active         & 0.62 ± 0.01 & 0.52 ± 0.04 & 0.54 ± 0.01 & 0.40 ± 0.02 & 0.27 ± 0.03 & 0.42 ± 0.08 \\
    Flappy Passive       & 0.67 ± 0.01 & 0.44 ± 0.04 & 0.52 ± 0.05 & 0.35 ± 0.06 & 0.26 ± 0.09 & 0.39 ± 0.12 \\
    Flappy Active        & 0.66 ± 0.06 & 0.46 ± 0.02 & 0.57 ± 0.03 & 0.51 ± 0.01 & 0.31 ± 0.03 & 0.51 ± 0.05 \\
    \bottomrule
\end{tabular}
\caption{Performance Across Multi-Subject Models (MLP): Average classifier performance (F1) for various levels of Model Granularity and Conditions. Binary and multi-class cross-validation saw mostly insignificant performance, but fine-tuning these models increased binary and multi-class model classification by nearly 17\% and 41\%, respectively. Random chance F1 performance was 0.50 for binary and 0.33 for multi-class classification.}
\label{tab:model_performance}
\vspace{-10pt}
\end{table*}
%
Figure \ref{fig:heatmap} shows three heatmaps illustrating multi-policy action agreement over three episodes for each optimality class. Each performance category is shown to be distinct to our system and a \textit{point of failure} can be identified within an episode.
\\
\textbf{Pre-Processing and Feature Extraction:}
We train a Support-Vector Machine (SVM), K-Nearest Neighbors (KNN), Random Forest, and Multilayer Perceptron (MLP) using a time series classification approach. Each model was trained and tested on a set of participant data for some condition(s). We used a custom dual-slope frequency-domain fNIRS probe to suppress superficial and motion artifacts. The raw fNIRS data was calibrated using the 20-second baseline at the beginning of each trial. Intensity and phase values recorded from the left and right PFC were measured at 690 nm and 830 nm with 110 MHz modulation, and converted to oxy-/deoxy-hemoglobin. Resulting features were sampled at 5.2 Hz and band-pass filtered (0.001 - 0.2 Hz; 3rd order).
\\
\textbf{Time Classification Approach:} We use a sliding window approach that extracts fixed-duration time windows, where each window was assigned a single label. In our experiments, window length varied per condition and was chosen through parameter studies. They were generally 5 to 7 seconds in length and 1 to 2 seconds in stride to address the $5$ to $7$ second latency of the fNIRS signal. The label assigned to an entire window was designated based on the endpoint label.
\\
We denote our time series as: $x^{k,p}_{1:T} = [x^{k,p}_{1}, x^{k,p}_{2}, \dots, x^{k,p}_{T}]
$ where each subject $k$ has neural data of length $T$, each terminating at an endpoint $p$. Each vector $x^{k,p}_t \in \mathbb{R}^F$ represents the F-dimensional data at time $t$. The endpoint $p$ is associated with a label, as discussed above. Slope, mean, standard deviation, intercept, skewness, and kurtosis were calculated for each feature window resulting in a vector of length $F = 8 \cdot 6$.
\\
\textbf{Training Paradigms:}
Our models were trained and evaluated using one of three different paradigms: single-subject, multi-subject and fine-tuned training. These paradigms are based on prior work that aimed to better calibrate BCI models for specific users \cite{huangfNIRS2MW2021}. As illustrated in \ref{fig:paradigms}, a single-subject model is trained on data from one participant and is validated using withheld data from the same participant. Multi-subject models were trained on data from a set of participants and evaluated using withheld data from the same set. Fine-tuned models are multi-subject models that were further trained with a fraction of a target participant's data and evaluated using withheld data from that target participant. Multi-subject and fine-tuned training data was exclusive to one condition, or a set of passive or active conditions.

To adjust for an imbalanced class distribution, we randomly down-sampled the majority class. We use a 60-20-20 train-test-validation split. Fine-tuned models use 20\% of the target participant's down-sampled data to train on, and the remaining data was reserved for testing.
%
%
%
\begin{figure*}[ht]
\includegraphics[width=16cm]{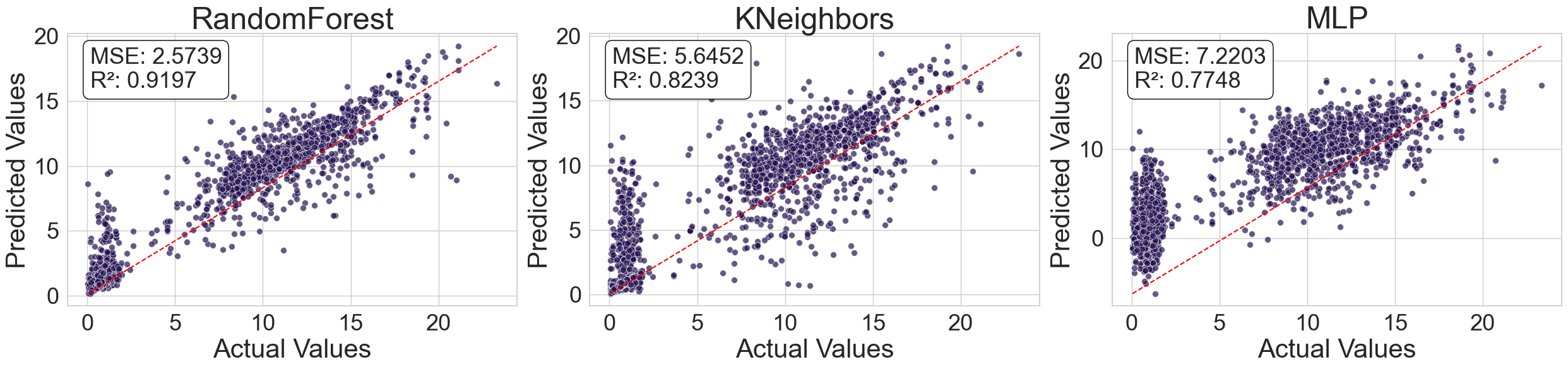}
\centering
\caption{Regression for Multi-Subject Robot Passive Task: This figure compares actual and predicted performance values for participants passively observing the Robot condition. The model uses fNIRS data to predict a continuous performance label. All models showed strong correlation between fNIRS signals and agent performance, with MSE, $R^2$ values, and corresponding scatter plots.}
\label{fig:regressor}
\end{figure*}
\\
\textbf{Cross-Validation Studies:} 
leave-one-subject-out (LOPO), and in some cases leave-two-subjects-out, approach. From the set of participants in a condition set, one or two of those participants were completely withheld as a cross-validation testing group. These participants were excluded from the training set, allowing cross-subject evaluation of the model within the same condition.

%% file: Sections/results.tex
\section{Results}
Our study evaluated the performance of binary and multi-class classification models across various task conditions, alongside cross-validation studies and a regression analysis. Below, we summarize our key findings.
\\
\textbf{Single-Subject Models:}  Single-subject models successfully distinguished between binary ($F1= 0.79$) and multi-class ($F1= 0.75$) levels of agent performance. Multi-class classifiers successfully differentiated between optimal, suboptimal, and worst-case behavior with little variance. However, cross-validation studies revealed that no single-subject model was able to predict agent performance from the neural data of another subject.
\\
\textbf{Multi-Subject Models:}  Binary multi-subject classifiers performed best on average, effectively learning patterns across a set of users in a specific condition, or set of conditions. Multi-class classification also showed promise, exceeding random chance in all but one condition: Flappy Passive. Like the Flappy Passive condition, some multi-class models struggled to distinguish between adjacent classes of agent performance. Task-specific challenges and variations in participant attention may have influenced performance, as the Flappy Active multi-class model performed much better than its Passive counterpart. Although model performance was generally consistent across conditions, Lunar Lander and Flappy Bird conditions showed higher variance and lower performance compared to both Robot conditions, shown in Table \ref{tab:model_performance}.
\\
\textbf{Cross-Validation Studies:} Cross-validation studies revealed little success across conditions and participants. Only the Robot Passive condition demonstrated any notable potential for zero-shot transferability when evaluated on users within the same condition ($F1 = 0.54$, $\sigma = 0.01$). This finding aligns with prior work and suggests that developing generalizable BCI models, as explored in \cite{huangfNIRS2MW2021}, remains an open challenge. These findings motivate the application of domain adaptation and deep learning techniques to improve robustness across participants and tasks~\cite{wang_taming_2021,eastmond_deep_2022}.
\\
\textbf{Fine-tuned Models:}  Fine-tuned models leverage a limited amount of target participant data to calibrate multi-subject models, leading to improved cross-validation performance. With only about 20\% of the downsampled target participant data, average scores increased by 16.9\% and 41.3\% for binary and multi-class models, respectively. These results suggest that even a small fraction of participant-specific data can help calibrate the model, boosting performance above random chance for an unseen, target participant.
\\
\textbf{Regression Performance:}  Most regression models performed particularly well, highlighting the ability to extract deeper complexity from fNIRS data beyond scalar feedback. Figure \ref{fig:regressor} illustrates the performance of MLP, K-Nearest Neighbors, and Random Forest regression models in the Robot Passive condition. Regressors trained on Active conditions performed marginally better on average ($R^2 = 0.81$) than Passive conditions ($R^2 = 0.77$). However, these models were unable to generalize to new participants. Finetuning with 30\% of the target participant's data increased these scores, but did not raise performance to a sufficient degree for application in any condition other than Lunar Active (Cross-Sub $R^2 = 0.01$, Fine-tuned $R^2 = 0.23$). Regression models may need more target data to calibrate multi-subject models.
\\
\textbf{NASA-TLX Results:} NASA-TLX allows participants to self-report mental workload through six different lenses: Mental, Temporal, Performance, Frustration, Physical, and overall Effort Demands. Self-reported cognitive load during Active tasks was higher for nearly every category (Figure \ref{fig:nasa-tlx}). Lunar Lander Active tasks were particularly high across all categories. Passive tasks were least demanding; participants verbally described watching the autonomous Lunar Lander and Flappy Bird agent as ``boring''. These experiences may contribute to slightly lower Passive condition performance.

%% file: Sections/conclusion.tex
\section{Discussion}
Our results indicate that both binary and multi-class classifiers are capable of identifying performance categories with meaningful accuracy, and regression analysis further reveals that neural responses contain fine-grained evaluative signals beyond categorical labels. Our cross-subject evaluations highlight promising potential for generalization in the Robot Passive condition, suggesting that task structure and participant engagement may play critical roles in model transferability. 

We identified three key limitations in this work. First, our dataset was partially imbalanced: some condition episodes were shorter, leading to fewer samples and underrepresented classes. While we addressed this via downsampling, the reduced data volume likely limited generalization and performance in these conditions. Second, self-reported NASA-TLX and post-task questionnaires revealed lower engagement and cognitive effort in certain passive conditions, which may have affected neural signal quality. Finally, while functional near-infrared spectroscopy (fNIRS) offers a non-invasive and accessible modality for Brain-Computer Interfaces (BCIs), it remains subject to signal noise, motion artifacts, and inter-subject variability. These technical challenges highlight the need for continued development of robust preprocessing techniques for neural classification tasks.

A natural next step would be to integrate these models into real-time RLHF frameworks to enable adaptive agent behavior based on implicit neural feedback. Future work should also explore personalized calibration strategies, and take advantage of multi-modal feedback (e.g., EEG, EMG, GSR) for greater insights into internal human assessments. Finally, we highlight the need to expand datasets to further support cross-subject generalization.
\begin{figure}[b!]
\includegraphics[width=8cm]{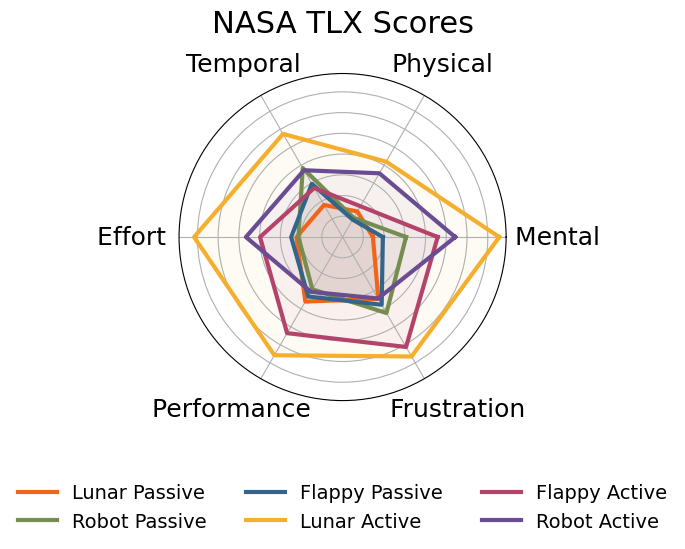}
\centering
\caption{NASA-TLX Scores per Condition: NASA-TLX is a self-reported questionnaire that measures a participant's perceived cognitive workload. Average scores showed greater perceived cognitive workload for active tasks. Lunar Passive and Flappy Passive tasks were lowest overall.}
\label{fig:nasa-tlx}
\end{figure}
\section{Conclusion}
This paper demonstrates a measurable relationship between fNIRS data and agent performance, showing that fNIRS signals can be decoded into meaningful evaluative feedback. We trained models to distinguish levels of agent performance based solely on passive neural responses, and repeated our procedures using active demonstration to serve as a benchmark against explicit feedback modalities. We show that passive fNIRS signals can be mapped to various levels of agent performance, laying the groundwork for enabling seamless integration of neural feedback into RLHF systems.